\documentclass{article}

\usepackage{microtype}
\usepackage{graphicx}
\usepackage{subfigure}
\usepackage{booktabs} 

\usepackage{hyperref}

\usepackage[accepted]{icml2024}

\usepackage{amsmath}
\usepackage{amssymb}
\usepackage{mathtools}
\usepackage{amsthm}

\usepackage[capitalize,noabbrev]{cleveref}

\usepackage{enumitem}
\usepackage{titlesec}
\usepackage{graphicx}
\usepackage{caption}
\usepackage[font={small}]{caption}

\theoremstyle{plain}

\theoremstyle{definition}

\theoremstyle{remark}

\usepackage[textsize=tiny]{todonotes}

\icmltitlerunning{
Video as the New Language for Real-World Decision Making
}

\begin{document}

\twocolumn[
\icmltitle{
Video as the New Language for Real-World Decision Making
}



\icmlsetsymbol{equal}{*}

\begin{icmlauthorlist}
\icmlauthor{Sherry Yang}{comp,yyy}
\icmlauthor{Jacob Walker}{comp}
\icmlauthor{Jack Parker-Holder}{comp}
\icmlauthor{Yilun Du}{sch}
\icmlauthor{Jake Bruce}{comp}
\icmlauthor{Andre Barreto}{comp}\\
\icmlauthor{Pieter Abbeel}{yyy}
\icmlauthor{Dale Schuurmans}{comp}
\end{icmlauthorlist}

\icmlaffiliation{comp}{Google DeepMind}
\icmlaffiliation{yyy}{UC Berkeley}
\icmlaffiliation{sch}{MIT}

\icmlcorrespondingauthor{Sherry Yang}{sherryy@\{google.com\}\{berkeley.edu\}}

\icmlkeywords{Machine Learning, ICML}

\vskip 0.3in
]



\printAffiliationsAndNotice{} 

\begin{abstract}

Both text and video data are abundant on the internet and support large-scale self-supervised learning through next token or frame prediction. However, they have not been equally leveraged: language models have had significant real-world impact,
whereas video generation has remained largely limited to media entertainment.
Yet video data captures important information about the physical world that is difficult to express in language.
To address this gap, we discuss an under-appreciated opportunity to extend video generation to solve tasks in the real world. 
We observe how, akin to language, video can serve as a unified interface that can absorb internet knowledge and represent diverse tasks. 
Moreover, we demonstrate how, like
language models, video generation can serve as planners, agents, compute engines, and environment simulators through techniques such as in-context learning, planning and reinforcement learning. 
We identify major impact opportunities in domains such as robotics, self-driving, and science, supported by recent work that demonstrates how such advanced capabilities in video generation are plausibly within reach.
Lastly, we identify key challenges in video generation that mitigate progress.
Addressing these challenges will enable video generation models to demonstrate unique value alongside language models in a wider array of AI applications.

\end{abstract}



\setlength{\textfloatsep}{1ex}

\vspace{-5mm}
\section{Introduction}

There has been tremendous progress in training large language models (LLMs) from internet text datasets in the past few years \citep{team2023gemini,achiam2023gpt}. The impressive performance of LLMs on a wide variety of tasks makes it tempting to reduce the artificial intelligence agenda to scaling up these systems. However, this is not sufficient. Firstly, the quantity of publicly available text data is becoming a bottleneck to further scaling~\citep{villalobos2022will}. Secondly, and perhaps more importantly, natural language alone might not be enough to describe all intelligent behavior~\citep{searle1980minds,dennett1993consciousness,minsky1988society} or capture all information about the physical world we live in (e.g., imagine teaching someone how to tie a knot using words only). While language is a powerful tool to describe higher-level abstractions, it is not always sufficient to capture the physical world in all its wealth of detail. 

Thankfully, there are abundant video data on the internet (e.g., over ten thousand years of consecutive video watching from YouTube alone) encapsulating a wealth of information imbued with knowledge of the world. Nevertheless, today's machine learning models trained on internet text or video data have demonstrated remarkably different capabilities. LLMs have advanced to tackling intricate tasks that require sophisticated reasoning~\citep{huang2022towards}, tool use~\citep{mialon2023augmented}, and decision making~\citep{yang2023foundation}. In contrast, video generation models have been less explored, primarily focusing on creating entertainment videos for human consumption~\citep{ho2022imagen,singer2022make,bar2024lumiere}. Given the paradigm shift unfolding in language modeling, it is important to ask whether we can elevate video generation models to the level of autonomous agents, simulation environments, and computational engines similar to language models so that applications requiring visual modalities such as robotics, self-driving, and science can more directly benefit from internet visual knowledge and pretrained video models.

In this paper, we take the position that \emph{video generation will be to the physical world as language modeling is to the digital world.} To arrive at this position, we first identify key components that have enabled language models to solve many real-world tasks: (1) a \emph{unified representation} (i.e., text) that can absorb broad information from the internet, (2) a \emph{unified interface} (i.e., text generation) through which diverse tasks can be expressed as generative modeling, and (3) language models' ability to \emph{interact} with external environments (e.g., humans, tools, and other models) by taking actions and optimizing decisions based on external feedback through techniques such as reinforcement learning from human feedback~\citep{ouyang2022training}, planning~\citep{huang2022language}, search~\citep{yao2023tree}, and optimization~\citep{rafailov2023direct}.

Motivated by these three aspects of language models, we observe that (1) video can serve as a unified representation to absorb broad information about the physical world, (2) diverse tasks in computer vision, embodied AI, and science can be expressed or supported by a video generation model, and (3) video generation as a pretraining objective introduces internet-scale supervision for large vision models, behavior models, and world models, which in tern enables actions to be extracted, environment interactions to be simulated, and decisions to be optimized.

To further illustrate how video generation can have a profound impact on real-world applications, we provide an in depth analysis on recent work that utilizes video generation as task solvers, answers to questions, policies/agents, and environment simulators through techniques such as instruction tuning, in-context learning, planning, and reinforcement learning (RL) in settings such as games, robotics, self-driving, and science. Lastly, we identify major difficulties around video generation, and suggest plausible solutions to address these challenges to unleash the full potential of video generation in the real world.


\section{Preliminaries}

We provide a brief overview of video generation models and how they have been used in domain-specific settings through conditional generation.

\subsection{Conditional Video Generation}
We denote a video clip as a sequence of image frames $\textbf{x} = (x_0, ..., x_t)$. An image on its own can be treated as a special video with a single frame, $\textbf{x} = (x_0,)$. Conditional video generation models the conditional probability $p(\textbf{x}|c)$ where $c$ is the conditioning variable. The conditional probability $p(\textbf{x}|c)$ has commonly been factorized by an autoregressive model~\citep{razavi2019generating}, a diffusion model~\citep{ho2022imagen}, or a masked transformer model~\citep{chang2022maskgit}. Depending on the factorization, sampling from $p(\textbf{x}|c)$ corresponds to either predicting images (patches) sequentially or predicting all frames $(x_0, ..., x_t)$ together, iteratively.

\subsection{Task-Specific Specialization}
Depending on what is in the conditioning variable $c$, conditional video generation can serve different purposes. Below, we enumerate common examples of $c$ and their use cases.

\begin{itemize}[leftmargin=*, topsep=0pt, itemsep=0pt, parsep=0pt]
    \item $p(\textbf{x}|c=\text{text})$. This corresponds to text-to-video models commonly used for generative media~\citep{kondratyuk2023videopoet,blattmann2023align}, where the text is often some creative description of the desired video (e.g., ``A teddy bear painting a portrait'' in \citet{singer2022make}). Text-to-video has mostly been applied to generating movies~\citep{zhu2023moviefactory} and animations~\citep{he2023animate,guo2023animatediff}.
    
    \item $p(\textbf{x}|c = \{x_0, \text{text}\})$. This corresponds to generating video rollouts starting from a given image $x_0$ while incorporating the text description. This type of conditioning has been applied to generate scene-specific visual interactions~\citep{yang2023learning} and visual plans for robot executions~\citep{du2023learning}. When $\textbf{x}$ only contains a future image $x_t$, $p(x_t|c = \{x_0,\text{text}\})$ can predict visual goals for robot manipulation~\citep{black2023zero,yu2023scaling}. This approach to goal synthesis is largely inspired by the vast literature on stylized image generation and inpainting~\citep{efros2023image,wang2023imagen}.
    
    \item $p(\textbf{x}|c = \{\overline{\textbf{x}}, \text{text}\})$. When $\textbf{x}$ and $\overline{\textbf{x}}$ have the same underlying content, this corresponds to text-guided video editing and stylization~\citep{loeschcke2022text,yang2023probabilistic}, which has been applied to generate self-driving videos in different weather conditions~\citep{hu2023gaia}. Note that  $\overline{\textbf{x}}$ can also be completely different from $\textbf{x}$, in which case $\overline{\textbf{x}}$ can serve as a visual prompt to elicit certain patterns in the output videos~\citep{bai2023sequential}.
    
    \item $p(x_{i + 1}|c =\{x_i, \text{action}\})$. This corresponds to learning a visual dynamics model where the action can be robot controls~\citep{yang2023learning}, keyboard inputs~\citep{hafner2020mastering}, or other motion information~\citep{li2023generative} that causes change in the visual space. If we replace $x_{i+1}$ with $x_{i+t}$ for some $t > 1$, we have a temporally-abstract dynamics model~\citep{sutton99between}. In this case we can also replace $x_i$ with any sub-sequence of $(x_i, x_{i+1}, ..., x_{i +t-1})$.
\end{itemize}

These specializations of conditional video generation suggest that there may exist a general framework under which broad video data can be absorbed and diverse tasks can be expressed using video generation.

\section{Unified Representation and Task Interface}

In this section, we first describe how video is a \emph{unified representation} that can capture various types of information from the internet to form broad knowledge. We then discuss how diverse tasks from computer vision and embodied AI can be formulated as a conditional video generation problem, providing the foundation for real-world decision making with video generation. Details of the models used to generate the examples can be found in Appendix~\ref{app:model}. Additional generated videos can be found in Appendix~\ref{app:result}.

\subsection{Video as a Unified Representation of Information}
\label{sec:unified-info}

While internet text data has provided much value to the digital/intellectual world with large language models, text is more suitable for capturing high-level abstractions as opposed to low-level details of the physical world. Below, we list a few types of information that is hard to express as text but can be easily captured by video.

\begin{itemize}[leftmargin=*, topsep=0pt, itemsep=0pt, parsep=0pt]
    \item \textbf{Visual and Spatial Information}: This includes visual details such as colors, shapes, textures, lighting effects, and spacial details such as how objects are arranged in space, their relative positions, distances, orientations, and 3D information. Such information naturally exist in image/video format as opposed to text format.
    \item \textbf{Physics and Dynamics}: This includes details about how objects and environments interact with each other physically, such as collisions, manipulations, and other movements influenced by physical laws. While text can describe movements at a high-level (e.g., ``a car driving down the street''), it is often insufficient to capture low-level details such as the torque and friction applied to the vehicle. Videos can implicitly capture this information.
    \item \textbf{Behavior and Action Information}: This includes information such as human behaviors and agent actions, characterizing the low-level details of performing tasks such as how to assemble a piece of furniture. Text again can mostly capture high-level descriptions of how to perform a task as opposed to the detailed information such as precise motions and movements.
\end{itemize}

\paragraph{Why Video?} One may wonder that, even if text is not sufficient to capture the above information, why video? To answer this question, we observe that video, in addition to existing at an internet scale, is interpretable to humans (similar to text) so that debugging, interaction, and safety speculation can be easily conducted. Moreover, video is a flexible representation that can characterize information at different spacial and temporal resolutions, e.g., atoms moving at angstrom scale ($10^{-10}$m)~\citep{kashin2021neural} and light traveling at a trillion frames per second~\citep{faccio2018trillion}.

\subsection{Video Generation as a Unified Task Interface}
\label{sec:unified-task}

In addition to a unified representation that can absorb broad information, we have seen from language modeling that we need a unified task interface through which diverse tasks can be expressed using a single objective (e.g., next token prediction); also, it is the alignment between representation of information (e.g., text) and task interface (e.g., text generation) that enables transfer of broad knowledge to task-specific decisions. In this section, we show how diverse vision tasks, as well as a broader set of question answering, reasoning, and problem solving, can all be expressed as a video generation task.

\begin{figure}[t]
    \centering
    \includegraphics[width=\linewidth]{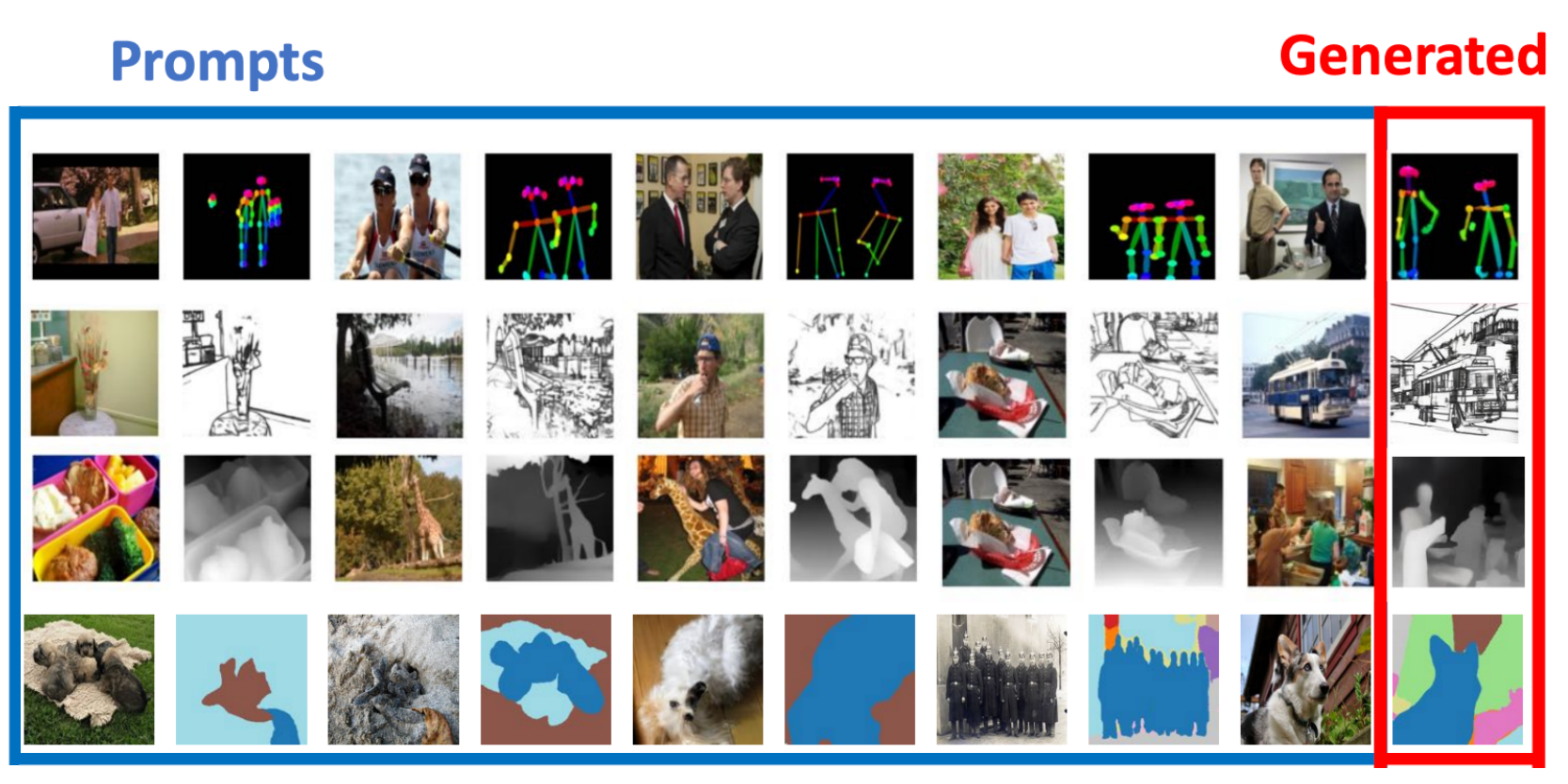}
    \caption{\textbf{Vision Tasks as Video Generation.} Figure 8 from~\citet{bai2023sequential} (simplified to show partial prompts), where diverse computer vision tasks such as joint/edge detection, depth estimation, and segmentation can be converted into a single next-frame prediction task.}
    \label{fig:lvm}
\end{figure}

\paragraph{Classical Computer Vision Tasks.} 
In natural language processing, many tasks (e.g., machine translation, text summarization, question answering, sentiment analysis, named entity recognition, part-of-speech tagging, text classification, dialogue systems) have traditionally been considered as different tasks but now have all been unified under the umbrella of language modeling. This has allowed greater generalization and knowledge sharing across tasks. Similarly, computer vision also has a broad set of tasks spanning across semantic segmentation, depth estimation, surface normal estimation, pose estimation, edge detection, and object tracking. Recent work has shown that it is possible to convert diverse vision tasks into a video generation task as shown in Figure~\ref{fig:lvm}~\citep{bai2023sequential,bar2022visual,wang2023images}, and that this unified approach to solving vision tasks scale favorably with model size, data size, and context length~\citep{bai2023sequential}.

Converting vision tasks into a video generation task generally involves the following steps:  (1) structure the input and output of a task (e.g., segmentation maps, depth maps) into a unified image/video space, (2) reorder image frames so that an input image is followed by the expected output image of a specific task (e.g., a regular input image followed by a depth map), and (3) leverage in-context learning by providing example input-output pairs as input to the conditional video generation model to specify the desired task.

\begin{figure}[t]
    \centering
    \includegraphics[width=\linewidth]{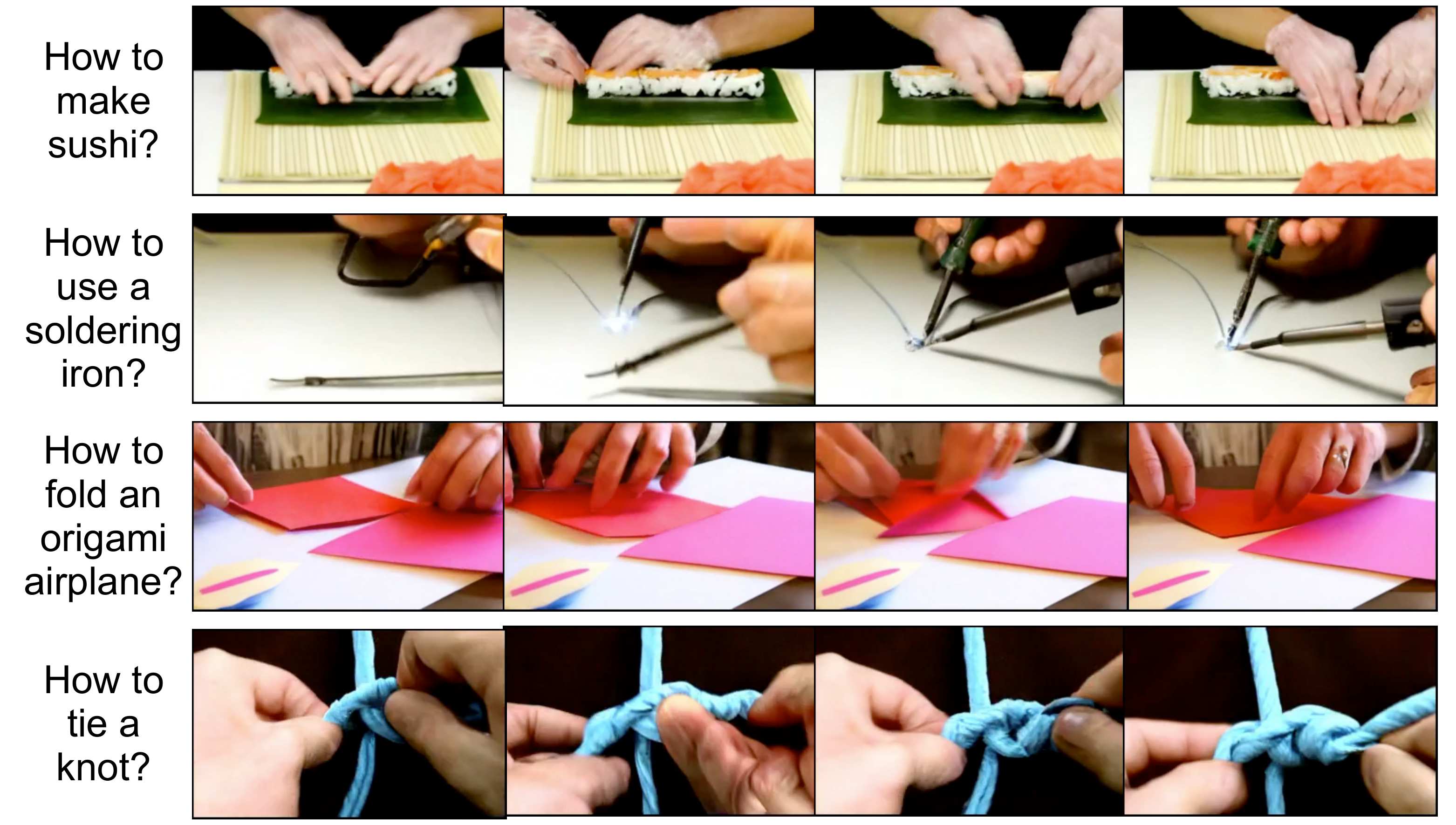}
    \caption{\textbf{Generated How-to Videos.}  Video generation models can synthesize key frames corresponding to human hands performing intricate tasks. However, the generated frames are too generic and do not capture enough details to fully answer users' questions.}
    \label{fig:howto}
\end{figure}

\paragraph{Video as Answers.} In traditional visual question answering (VQA)~\citep{antol2015vqa}, the expected answers are in text. With the development in video generation, a novel task would be to treat video as answers, e.g., a video would be generated in response to ``how to make an origami airplane''~\citep{souvcek2023genhowto,yang2023learning}. Similar to how language models can generate customized response to human inquiries in text, video models can also generation customized answers to how-to questions with great low-level details. Such video response can be more preferable to humans than textual response~\cite{yadav2011if}. In Figure~\ref{fig:howto}, we illustrate videos generated by a text-to-video model in response to a set of how-to inquiries. Additionally, one may consider conditioning generation on an initial frame to synthesize video answers in user-specific scenes. Despite such a grand promise, videos synthesized by today's text-to-video models are generally too short/simple, not containing enough information to fully answer users' questions. 

The problem of synthesizing video frames to answer users' questions has similarities to planning with language models~\citep{valmeekam2023planbench}, except that both the state and low-level action spaces are now pixels as opposed to text. One may utilize language models or vision language models to break down high-level goals (e.g., ``how to make sushi'') into specific subgoals (e.g., ``first, put rice on rolling mat'') and synthesize plans for each subgoal while validating the plausibility of synthesized plans~\citep{du2023video}. 

\begin{figure}[b]
    \centering
    \includegraphics[width=\linewidth]{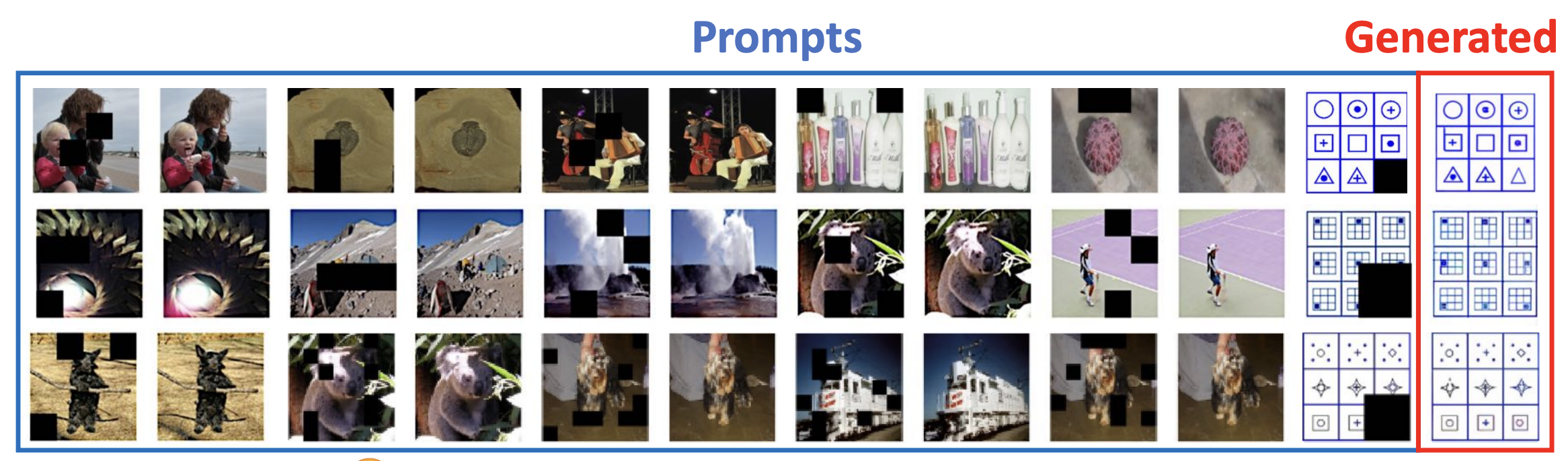}
    \caption{\textbf{Visual Reasoning as Next--Frame Generation.} Figure 13 from \citet{bai2023sequential} shows next-frame prediction can solve visual reasoning tasks such those in IQ tests.}
    \label{fig:reasoning}
\end{figure}

\paragraph{Visual Reasoning and Chain-of-Thought.} With a unified representation of information and a unified task interface, reasoning has emerged in language modeling where a model can elicit relevant information as intermediate steps towards solving more complex problems~\citep{wei2022chain}. Similarly, with video as a unified representation and task interface, video generation has also exhibited early signs of visual reasoning by predicting masked regions of an image, as shown in Figure~\ref{fig:reasoning}~\citep{bai2023sequential}. It would be interesting to see if next frame prediction can be used to solve more complex geometry problems similar to \citet{trinh2024solving} by generating videos with the right set of auxiliary lines.

\begin{figure}[t]
    \centering
    \includegraphics[width=\linewidth]{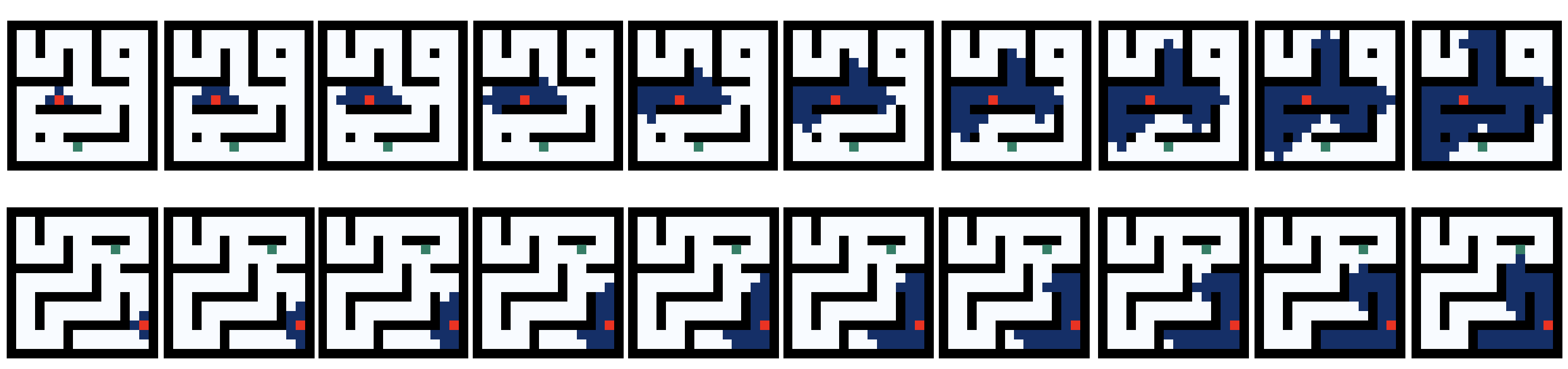}
    \caption{\textbf{BFS as Video Generation.} Figure 14 from \citet{yang2022chain} shows two sets of intermediate frames generated by a video model emulating the BFS search procedure. The red and green cells represent the start and goal locations. The white and black cells represent empty spaces and obstacles. Blue cells represent the cells that would have been visited by running the BFS algorithm.}
    \label{fig:cot}
\end{figure}

Building on the idea of leveraging next-frame prediction for visual reasoning and solving geometry problems, we can further characterize the reasoning \emph{process}~\citep{himakunthala2023let} and algorithms~\citep{yang2022chain} using videos. Specifically, \citet{yang2022chain} characterized the execution state of a Breadth First Search (BFS) algorithm using videos. In this context, learning to generate a video corresponds to learning to search, as illustrated in Figure~\ref{fig:cot} (see also \citet{silver2017predictron}). While the examples in Figure~\ref{fig:reasoning} and Figure~\ref{fig:cot} might seem contrived, they serve as early indicators that video generation as a pretraining task may elicit reasoning-like behaviors similar to language models, revealing opportunities in leveraging video generation to solve complex reasoning and algorithmic tasks.

\begin{figure*}[t]
    \centering
    \includegraphics[width=.9\linewidth]{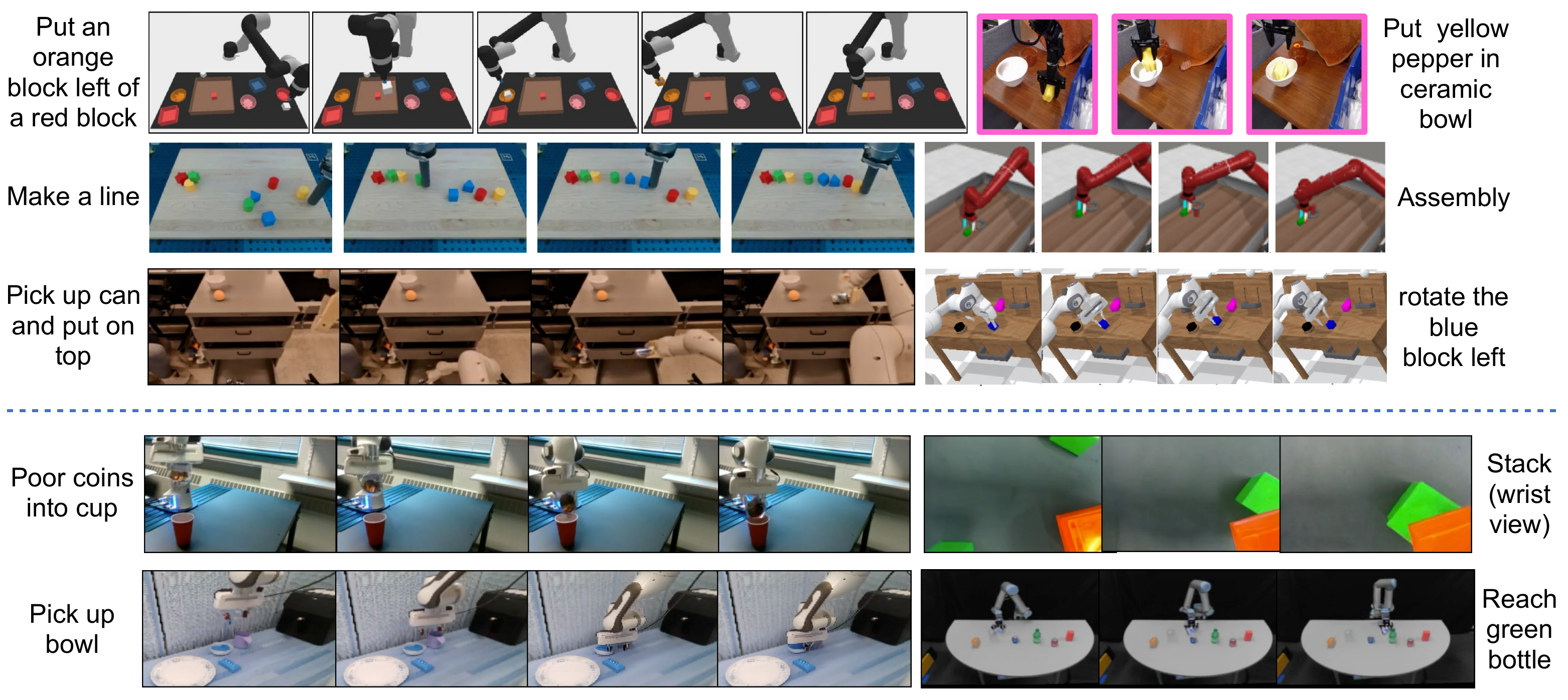}
    \caption{\textbf{Generated Video Plans for Robots.} [Top] Video plans generated by existing work (Figure 3 in~\citet{du2023learning}, Figure 3 in~\citet{black2023zero}, Figure 3 in~\citet{du2023video}, Figure 5 in~\citet{ko2023learning}, Figure 14 in~\citet{yang2023learning}, Figure 7 in~\citet{kang2023imagined}). [Bottom] Video plans generated by a single video generation model trained on the Open X-Embodiment~\citep{padalkar2023open}.}
    \label{fig:rtx}
    \vspace{-6mm}
\end{figure*}

\subsection{Video as a Unified State-Action Space}
\label{sec:state-action}

We have seen that video generation can absorb broad knowledge and characterize diverse vision tasks. In this section, we further support this observation by providing concrete examples in embodied AI of using video as a unified representation and task interface. 

One of the long-standing challenges in embodied AI has been data fragmentation, where datasets collected by one robot performing one set of tasks is hardly useful for learning on a different robot or on a different set of tasks~\citep{padalkar2023open}. The major difficulty in knowledge sharing across robots and tasks lies in that each type of robot and task has distinct state-action spaces. To address this difficulty, \citet{du2023learning} advocate the use of the pixel space as a unified state-action space across tasks and environments. Under this framework, embodied planning can be cast as a conditional video generation problem, thereby benefiting from internet pretrained video generation models.
An additional module such as an inverse dynamics model~\citep{du2023learning}, a goal-conditioned policy~\citep{black2023zero,kang2023imagined,du2023video}, an optical flow network~\citep{ko2023learning}, or dense grid point~\citep{wen2023any} can then be employed to recover the low-level robot controls from high-level video plans. We illustrate video plans generated by previous work in Figure~\ref{fig:rtx} (top). Most existing work trains one video generation model per robot, which diminishes the potential benefit of using video as a unified state-action space for embodied learning. We provide additional generated video plans from training a video generation model on the Open X-Embodiment dataset~\citep{padalkar2023open} with a diverse set of robots and tasks in Figure~\ref{fig:rtx} (bottom). Both the previous and newly generated video plans look highly realistic and successfully complete the specified tasks.

\section{Video Generation as Simulation}

While video generation on its own can already solve many tasks as described in the previous section, another important opportunity in video generation is to simulate visual observations of various systems and processes, so that control inputs to a system can be optimized according to simulation results. This is especially useful for applications where abundant video data can be collected but the underlying dynamics are hard to be explicitly expressed (e.g., cloud movement, interaction with soft objects). In this section, we begin by studying such visual generative simulators in game settings, where we may have ground truth game engines to verify qualities of learned simulators and iterate on effective generation of new experiences. We then provide examples of simulating real-world processes such as robot interactions, autonomous driving, and atomic-level interactions. Details of the generative models used to generate the examples can be found in Appendix~\ref{app:model}. Additional generative simulation results can be found in Appendix~\ref{app:result}.

\subsection{Generative Game Environments}
Games have been used as a testbed for AI algorithms for decades \citep{yannakakis2018artificial}. For instance, the Arcade Learning Environment~\citep{bellemare2013arcade} enabled the development of deep Q-learning, the first AI agent to reach human level in playing Atari games~\citep{mnih2015human}. In a similar vein, we can consider games as a means to test the quality of generative simulators by comparing against ground truth simulations from the game engine. In the future, we may even be able to use generative models to surpass what is possible with existing human-designed simulated environments. In this section we discuss these possibilities, ranging from simulating a single complex environment to generating entirely new ones.

\paragraph{Simulating Complex Games.} \label{sec:minecraft}

\begin{figure}[t]
    \centering
    \setlength{\tabcolsep}{-1pt}
    \begin{tabular}{cccc}
     \includegraphics[width=.2\linewidth]{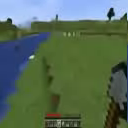} &
     \includegraphics[width=.2\linewidth]{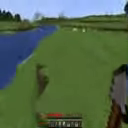} &
     \includegraphics[width=.2\linewidth]{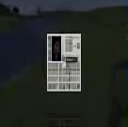} & 
     \includegraphics[width=.2\linewidth]{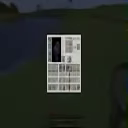} \\ 
     \includegraphics[width=.2\linewidth]{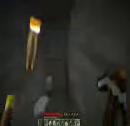} &
     \includegraphics[width=.2\linewidth]{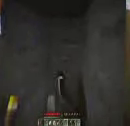} &
     \includegraphics[width=.2\linewidth]{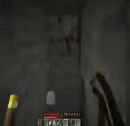} & 
     \includegraphics[width=.2\linewidth]{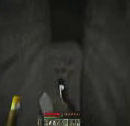} \\ 
     \includegraphics[width=.2\linewidth]{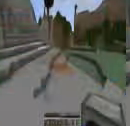} &
     \includegraphics[width=.2\linewidth]{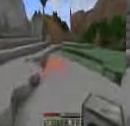} &
     \includegraphics[width=.2\linewidth]{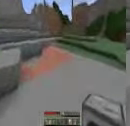} & 
     \includegraphics[width=.2\linewidth]{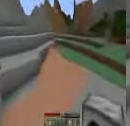} \\ 
    \end{tabular}
    \caption{\textbf{Generated Game Trajectories in Minecraft.} Both actions and observations are generated using an autoregressive model trained on Minecraft data. In the top row, the inventory is opened. The middle row shows use of a pickaxe to break a stone
    block. The bottom row predicts movement throughout the environment.}
    \label{fig:minecraft}
\end{figure}

Action-conditioned video generation can possibly simulate the environment dynamics of complex computer games such as Minecraft. As a proof of concept, we trained a transformer-based architecture, autoregressive in time, that predicts future agent actions and observations conditioned on episode history. We used the ``contractor data'' from~\citet{baker2022video}, which consists of trajectories collected while humans interacted with the game. Both observations and actions are quantized tokens, reducing model-based rollout to next token prediction. Note that in this case the model serves both as a world model and a policy: given a sequence of alternating observations and actions ending in an action, the model can infer the next observation (world model), and given an analogous sequence ending in an observation, the model can infer the next action to take (policy). Figure \ref{fig:minecraft} shows a few generated trajectories from this model. The model is capable of generating actions and transitions corresponding to sophisticated strategies (e.g., using a pickaxe to break a stone block).

With such a policy and dynamics backbone, model-based reinforcement learning algorithm---such as Dyna~\cite{sutton1991dyna}, Dreamer~\cite{hafner2020mastering}, and MuZero~\cite{schrittwieser2019mastering, antonoglou2022planning}---could be employed to improve the policy. This requires extensive sampling from the dynamics model, which in turn requires the generative model to be computationally efficient. Note that despite video generation models are highly general, when planning is concerned, world models perhaps do not have to be video models, and latent state space models have often been previously favored~\citep{ichter2019robot,hafner2020mastering}.

\paragraph{Generating Novel Game Environments.} 
Procedurally generating novel game contents and levels is an active area of research in the game AI community \citep{pcgml}, which has been shown to be useful to both training and evaluation of RL agents~\citep{pcg, pcg_illuminating, cobbe2020leveraging}. There have been attempts to leverage generative models for game design by directly predicting frames \citep{bamford2020nge} or modifying backgrounds to generate new game levels \citep{Kim2020_GameGan}. However, these works rely on privileged simulation data, and have only been attempted at a small scale, limiting the potential to generate entirely new game environments.

\begin{figure}[t]
    \centering
    \includegraphics[width=\linewidth]{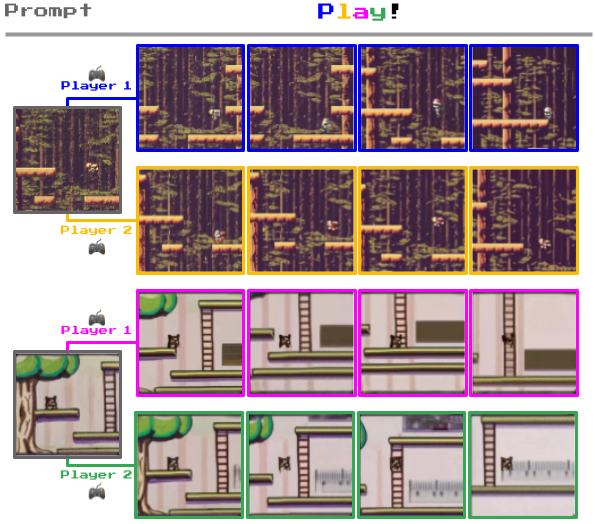}
    \caption{\textbf{Generated Interactive Game Environments}: Two synthetic image prompts passed to the model from \citep{genie2024}, which converts them into interactive environments. From there, it is possible to generate diverse trajectories by taking different latent actions, shown here as Player 1 and 2.}
    \label{fig:genie}
\end{figure}

Recent work has shown it is possible to leverage unlabelled internet-scale gameplay data to learn \emph{latent} actions and then train an action-controllable video model \citep{genie2024}. This makes it possible to generate an endless possibility of diverse interactive environments from a prompt image. Figure~\ref{fig:genie} shows generated game trajectories controlled by human players selecting latent actions given two novel starting frames. While this work remains exploratory, one could imagine a future where it is possible to also integrate learned reward models \citep{chan2023visionlanguage, du2023rewards, escontrela2023video} to train RL agents in fully generative game environments.

\subsection{Robotics and Self-Driving.} 
\label{sec:robot}
\begin{figure}[t]
    \centering
    \includegraphics[width=\linewidth]{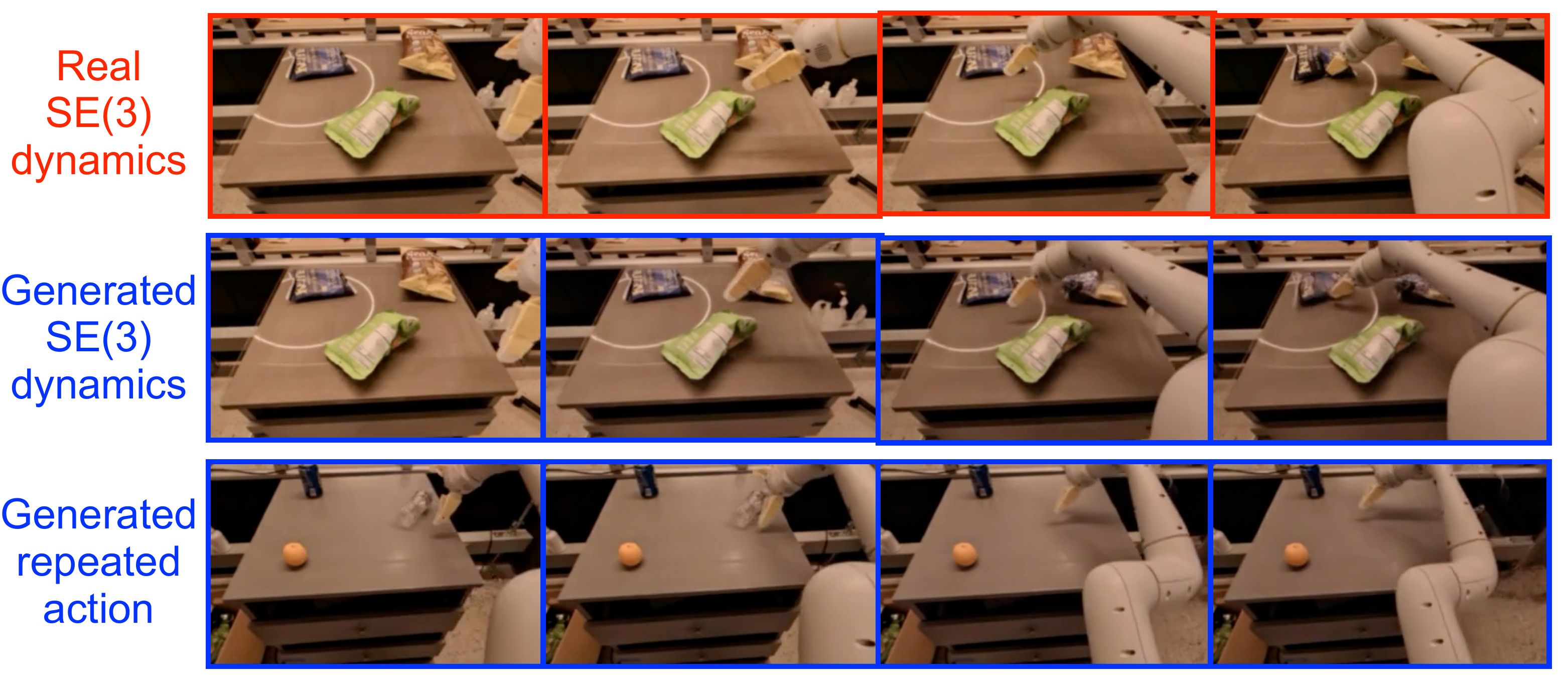}
    \caption{\textbf{Generative Simulation of SE(3) Robot Actions.} Real execution of a robot policy (top), simulated execution of the same policy (middle), and simulated  execution of repeating the same action (bottom). The simulated rollout generally agrees with the ground truth rollout, but hallucination can happen as the bottle disappears (bottom row).}
    \label{fig:dynamics}
\end{figure}

\paragraph{Simulating the SE(3) Action Space}

One of the long standing challenges in robot learning is around sim-to-real transfer~\citep{rusu2017sim}, where policies trained in a simulator fails to transfer to execution on a real robot.\citet{yang2023learning} demonstrated that it is possible to learn an action-conditioned next-frame prediction model on real-robot video data from the Language Table environment~\citep{lynch2023interactive} with a simple Cartesion action space. In Figure~\ref{fig:dynamics}, we illustrate that next-frame prediction can predict the visual effect of the more general end-effector actions in the SE(3) space~\citep{blanco2021tutorial}.

One immediate use case of a generative SE(3) simulator is to evaluate robot policies, which is particularly useful given the safety considerations associated with real-robot evaluation. Aside from evaluation, \citet{yang2023learning} has trained an RL policy using rollouts from a generative simulator in the Language Table environment. An interesting next step would be to learn a policy from both simulated rollouts and a real environment using a Dyna-style algorithm~\citep{sutton1991dyna}. Under this setting, real-world videos would be collected as the policy were being executed, which would serve as additional demonstration and feedback for the generative simulator. Lastly, generative simulators can enable effective training of multi-task and multi-environment policies through video rollouts in diverse environments. This was not possible previously, as a policy generally only has access to a single real-world environment at a time.

\begin{figure}[t]
    \centering
    \includegraphics[width=\linewidth]{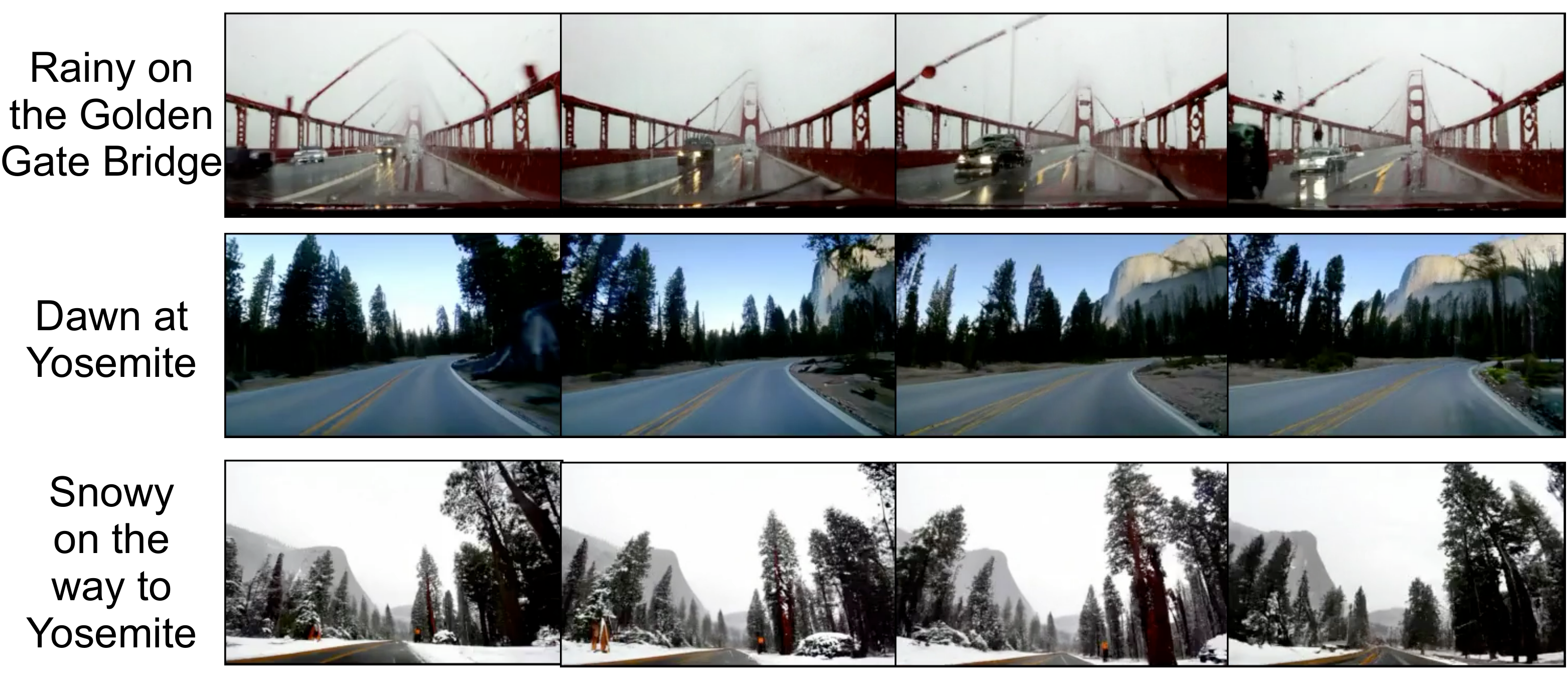}
    \caption{\textbf{Generative Simulation for Self-Driving.} With internet knowledge, we can simulate different driving conditions at particular locations, such as ``rain on Golden Gate Bridge'' (top), ``dawn in Yosemite'' (middle), and ``snow on the way to Yosemite'' (bottom).}
    \label{fig:driving}
\end{figure}

\paragraph{Domain Randomization.} Another benefit of generative simulators that is broadly applicable to robotics, navigation, and self-driving is their ability to introduce natural randomness to the training environment to improve real-world transfer of policies trained in simulation. Without generative models, this is achieved through domain randomization by hard-coding rendering rules~\citep{tobin2017domain}, which is tedious and results in limited environment variations and unrealistic rendering effects. With a generative simulator, recent work has shown that different driving conditions (e.g., sunny, foggy, snowy, rainy, at night) can be introduced into the simulator~\citep{hu2023gaia}. Furthermore, combined with internet-scale knowledge, we can simulate driving conditions at specific locations such as simulating driving in the rain on the Golden Gate Bridge, as shown in Figure~\ref{fig:driving}, which enables training self-driving policies with diverse locations and weather conditions.

\subsection{Science and Engineering}
\begin{figure}[t]
    \centering
    \includegraphics[width=\linewidth]{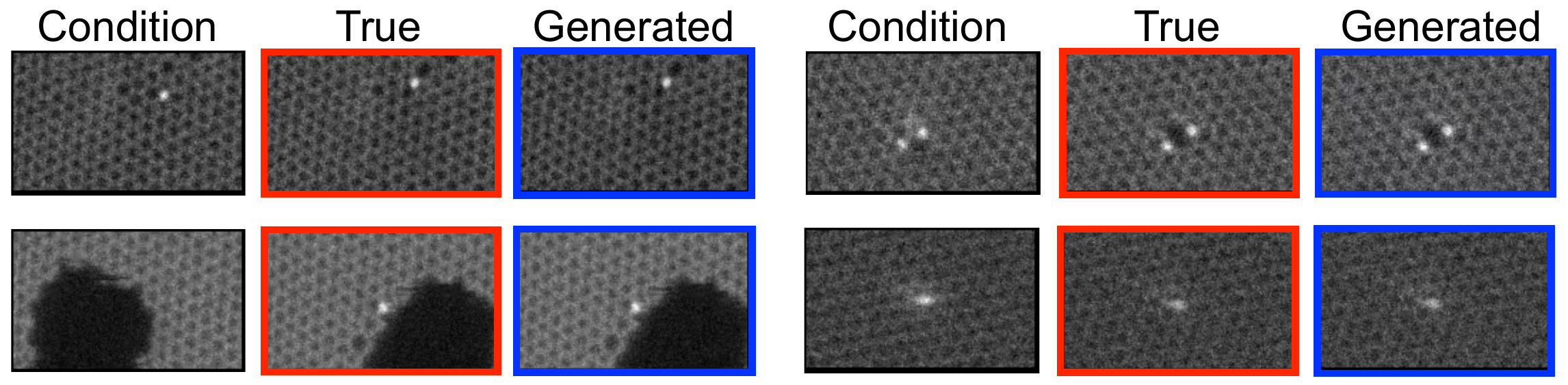}
    \caption{\textbf{Atomic-Level Next-Frame Prediction.} The conditional frame, true next frame, and generated next frame reflecting the visual dynamics of silicon atoms on graphene sheets stimulated by electron beams of an electron microscope. Generative models are capable of modeling the visual dynamics with high fidelity.}
    \label{fig:dune}
\end{figure}

Video can serve as a unified representation across a wide range of science and engineering domains, impacting research fields such as medical imaging, computerized image processing, and computational fluid dynamics~\citep{steinman2002image}. In situations where visual information can be easily captured by cameras but the underlying dynamical systems are difficult to identify (e.g., cloud movements, atom movements under electron microscopes), video generation models conditioned on the control input can be an effective visual simulator, which can then in tern be used to derive better control inputs. In Figure~\ref{fig:dune}, we illustrate the transition dynamics of silicon atoms on a single layer of carbon atoms, when stimulated by the electron beam of a scanning transmission electron microscope (STEM) using the STEM data collected from \citet{schwarzer2023learning}. We can see that the generative simulator is capable of characterizing the movement of the silicon atom in the pixel space.

Employing a highly realistic visual simulator in response to control inputs can mitigate the issue of limited hardware access in scientific research endeavors that requires operating specialized equipment, such as electron microscopes. However, leveraging a visual generative simulator for control input optimization requires further investigation to ensure its validity and effectiveness.

In addition to closing the sim-to-real gap in simulating scientific processes, another benefit of generative simulators is that they have a fixed computational overhead which can be beneficial when traditional computational methods are intractable. For instance, simulating calorimeter showers requires computing pairwise interactions between electrons, the complexity of which quickly becomes impractical when the number of electrons are large~\citep{mikuni2022score}. Videos of electron showers, on the otherhand, have a fixed computational overhaed in proportion to the resolution at which showers are being modeled.

\section{Challenges}
\label{sec:challenge}

While video generation has great potential, some major challenges for their application still remain. We outline these challenges and potential solutions in this section.

\subsection{Dataset Limitations}
\label{sec:challenge-data}

\paragraph{Limited Coverage.} 
In language modeling, the distribution of language data for solving specific downstream tasks is generally within the distribution of internet text data. However, this is not the case for video. Videos posted on the internet are geared towards human interest, which are not necessarily the video data useful for downstream tasks. For example, models for computational fluid dynamics would likely require many long videos focusing on the movement of fluids such as water; such videos lasting hours would not be very interesting to humans and are thus scarce on the internet. Similarly, it is unusual to find a particular type of robot (e.g., a Franka Emika Panda robot) performing a particular task (e.g., folding clothes) on the internet. This calls for better facilitation to collect and distribute domain specific video data. The Open-X Embodiment dataset for robotics is one such example~\citep{padalkar2023open}.

\paragraph{Limited Labels.} Another challenge in video modeling is the lack of annotated videos. For example, the MineDojo dataset~\cite{fan2022minedojo} has over 300 thousand hours of humans playing the game Minecraft, but the dataset only has language transcriptions but no game action labels, making it difficult to train policies or environment models using this dataset. Similarly, in the largest open-source robotics dataset~\citep{padalkar2023open}, many robot trajectories do not have language annotations on the tasks being performed, or only have generic labels such as ``interact with any object''.

In order to label more video data, prior work has utilized image/video captioning models to provide additional text labels which can be further used to train text-to-image/video models~\citep{betker2023improving,blattmann2023stable}. This is similar to video pretraining (VPT)~\citep{baker2022video} except that VPT labels video with action data as opposed to text data. Another possibility is to leverage latent actions/skills inferred from videos~\citep{edwards2019imitating,rybkin2018learning,ye2022become}, with the largest scale example being \citet{genie2024}. In Figure~\ref{fig:genie_additional_examples} in Appendix~\ref{app:result}, we show examples of the latent actions. Despite the consistency of learned latent actions, it remains an open question as to whether this approach could scale to more complex and diverse dynamics.

\subsection{Model Heterogeneity}

Unlike how language models have converged on an autoregressive architecture, video generation has yet to settle on the best approach. Autoregressive models, diffusion models, and masked models each have their own advantages and drawbacks.

\paragraph{Diffusion Models.} 
Diffusion models~\cite{sohl2015deep,ho2022imagen} (such as the model used in Section~\ref{sec:state-action}) have two major advantages. First, they can easily model continuous output spaces without requiring tokenization, which can lead to better generation quality. Second, multiple frames can be sampled in parallel. 
However, sampling speed in diffusion models is still fairly slow, limiting its applications in real-time simulation. In addition, it is unclear how to generate long video sequences with diffusion models. Diffusion models are also known to be sensitive to hyperparameters such as noise schedules~\citep{croitoru2023diffusion}, making training and scaling difficult.

\paragraph{Autoregressive Models.}
Autoregressive models with a tokenized output space (such as the model mentioned in Section~\ref{sec:minecraft}) are relatively easier to train than diffusion models. Tokenization also allows video generation to be integrated with text or discrete action generation, opening up more applications that require multi-modal generation~\citep{team2023gemini}. Additionally, autoregressive models scale well with context length~\cite{dai2019transformer, yan2023temporally,bai2023sequential}, allowing them to potentially model very long sequences of frames. However, autoregressive decoding is computationally expensive as each token has to be predicted sequentially. Furthermore, autogregressively bootstraped videos may suffer from the drifting effect~\citep{weng2023artboldsymbolcdotv}.

\paragraph{Masked Models.} Models based on masked reconstruction (such as the model used for generating novel game environments in Section~\ref{sec:minecraft}) can
leverage some of the benefits of diffusion and mitigate some
of the issues of token-autoregressive modelling
by sampling batches of image tokens in parallel~\cite{chang2022maskgit}.
This allows images
composed of thousands of tokens to be sampled with only dozens of
model invocations as in~\citet{genie2024}. However, this approach
introduces challenges such as sampling bias introduced by the
independence assumptions within individual sampling steps.

\paragraph{Better Future Models.} Potential solutions to model heterogeneity may require combining the advantages of different models
, such as combing autoregressive and masked models~\cite{yan2023temporally} or combining autoregressive and diffusion models~\citep{weng2023artboldsymbolcdotv}. In addition, video data might contain redundant information both spatially and temporally. Future models could consider learning latent spaces to reduce the redundency. Better video generation models should also address the current challenges in generation speed and long-term consistency across existing models.

\subsection{Hallucination}
\begin{figure}[t]
    \centering
    \includegraphics[width=\linewidth]{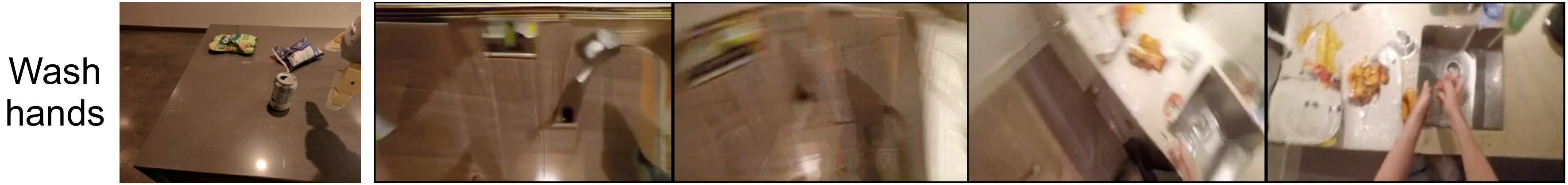}
    \caption{\textbf{Generation from an Unrealistic Instruction.} The input image to the video generation model is a table top with a robot hand. The language instruction is ``wash hand''. The video model is able to generate egocentric motions to move away from the table top to a kitchen sink in an attempt to fulfill the language instruction realistically.}
    \label{fig:unrealistic}
\end{figure}

Hallucination in video generation is common across different types of models. For instance, objects can randomly emerge or disappear (see Figure~\ref{fig:dynamics} bottom row and Appendix~\ref{app:hallucinate}). This could be due to the loss weight on objects often being not as high as the loss weight on backgrounds since objects are often small. Another type of common hallucination involves implausible dynamics, e.g., a cup ``jump'' into a robot hand as opposed to a robot grasping a cup. This could be due to videos with coarse temporal frequency not capturing the exact motion-critical frames. Furthermore, generative models that simultaneously model behaviors and dynamics may not distinguish visual changes caused by actions or dynamics~\citep{yang2022dichotomy}. Hallucination can also occur when a user input is unrealistic given a particular scene, e.g., ``wash hands'' is given to a table-top robot. Nevertheless, we haven seen that a video generation model attempts to generate realistic videos by utilizing egocentric motions to fulfill unrealistic user input as shown in Figure~\ref{fig:unrealistic}. Methods such as reinforcement learning with external feedback can be applied to further reduce hallucination in video generation models.

\subsection{Limited Generalization}

Generating videos from arbitrary image and text input has been difficult. This is especially true for domains that are not well represented by the training data, which, due to limited data coverage challenge discussed in Section~\ref{sec:challenge-data}, is quite common in practice. Take diffusion model as an example, it is a common to train on lower resolution videos followed by spatial super-resolution to prevent overfitting~\citep{ho2022imagen,bar2024lumiere,xing2023survey}. We hypothesize that high-resolution images/videos have too much high-frequency information invisible to human eyes, and the focus on which leads to a lack of generalization.

\section{Conclusion}
We have taken the position that video generation is to physical world as language modeling to the digital world. We have supported this position by showing how, similar to language models, video can represent broad information and tasks. We have further described prior work and new perspectives on applications of video generation combined with reasoning, in-context learning, search, planning, and reinforcement learning to solve real-world tasks. Challenges like hallucination and generalization notwithstanding, video generation models have the potential to become autonomous agents, planners, environment simulators, and compute engines, and to eventually serve as the artificial brain to think and act in the physical world.



\nocite{langley00}

\bibliography{example_paper}
\bibliographystyle{icml2024}

\newpage
\appendix
\onecolumn
\appendix
\clearpage
\begin{center}
{\huge Appendix}
\end{center}

\section{Details of Models Used to Generate Examples in the Main Text}
\label{app:model}

\subsection{Autoregressive Model}
\label{app:autoregressive}
The model in Section~\ref{sec:minecraft} is autoregressive in time but uses a masked model ~\cite{chang2022maskgit} for each frame in a manner similar to TECO~\cite{yan2023temporally}. For a given trajectory of pixel observation $x_t$ with corresponding action $a_t$, we model the interleaved sequence $z_0, a_0, z_1, a_1, ...$ where we encode pixel observations $x_t$ into tokens $z_t$ via VQVAE~\cite{van2017neural} in combination with a vision transformer~\cite{dosovitskiy2021an}. We tokenize the actions according to VPT~\cite{baker2022video}. We utilize Transformer-XL~\cite{dai2019transformer} to encode the temporal trajectory $z_0, a_0, z_1, a_1, ...$ with temporally aligned outputs $h_{z_0}, h_{a_0}, h_{z_1}, h_{a_1}, ...$. For steps where the last input was an observation, i.e. $h_{z_t}$, we utilize the context $h_{z_t}$ as conditioning input to an autoregressive transformer head to predict $a_t$. If the last input was an action, the context $h_{a_0}$ is conditioning to a masked transformer head to model $z_t$. Our MaskGIT implementation uses 8 steps with a cosine masking schedule. To further enhance the performance of our interleaved transformer, we initialize the memory using a \emph{past encoder}, an identical transformer separately trained on the interleaved sequence $..., x_{-2}, a_{-2}, x_{-1}, a_{-1}$ utilizing inputs without any discretization.

\subsection{Diffusion Model}
\label{app:diffusion}

The diffusion model used to generate examples in Figure~\ref{fig:howto}, Figure~\ref{fig:rtx}, Figure~\ref{fig:dynamics}, Figure~\ref{fig:driving}, and Figure~\ref{fig:dune} uses the same 3D U-Net architecture as ~\citet{ho2022video,ho2022imagen} with interleaved 3D attention and convolution layers in the spatial downsampling pass and spatial upsampling pass. Skip connections are applied to the downsampling pass activations. The model uses pixel-space diffusion as opposed to latent-space diffusion. Following conventions in video diffusion as described in Section~\ref{sec:challenge}, the lower resolution video generation model operates at resolution [24, 40], followed by two spacial super-resolution models with target resolution [48, 80] and [192, 320]. Classifier-free guidance~\citep{ho2022classifier} was applied for text or action conditioning. For frame conditioning, we input the conditioning frame into both the conditional and unconditional model used for classifier-free guidance. To simulate the SE(3) dynamics shown in Figure~\ref{fig:dynamics}, we employ action discretization similar to \citet{yang2023learning} and \citet{padalkar2023open}.

\subsection{Masked Model}
\label{app:masked}

The masked dynamics model in~\cite{genie2024} that generated the novel game environments in Section~\ref{sec:minecraft}
is a controllable
video continuation model, producing outputs autoregressively at the frame level,
conditioned on unsupervised latent variables that represent the transitions.
The latent variables are composed of a discrete set of VQ-VAE codes
~\cite{van2017neural} $\tilde{a}_{1:T-1}$ that are conditioned on frames $x_{1:T}$ and
optimized to help predict $\hat{x}_{2:T}$ with a causal transformer.
The dynamics model is a transformer with interleaved temporal and spatial attention
~\cite{gupta2022maskvit} trained using a masked
reconstruction objective, following~\cite{chang2022maskgit}. Video tokens are
masked with independent random Bernoulli masks at an average rate of 75\%, and
the dynamics model is trained to predict the missing tokens by minimizing a
cross-entropy objective.

At inference time, tokens are generated in parallel following
MaskGIT~\cite{chang2022maskgit}.
Beginning with unmasked context tokens for frames $x_{1:t-1}$ and a fully masked frame
$x_{t}$, a series of iterative steps are performed, where each step computes the
logits for all of the tokens conditioned on $x_{1:t}$ and $\tilde{a}_{1:t}$, a
candidate token for each remaining masked position is sampled, and the
highest-probability
samples are locked in for future steps. In~\cite{genie2024} each image is composed
of 920 tokens, and they are all eventually sampled over the course of 25 MaskGIT
steps.

\begin{figure}[h]
    \centering
    \includegraphics[width=.8\linewidth]{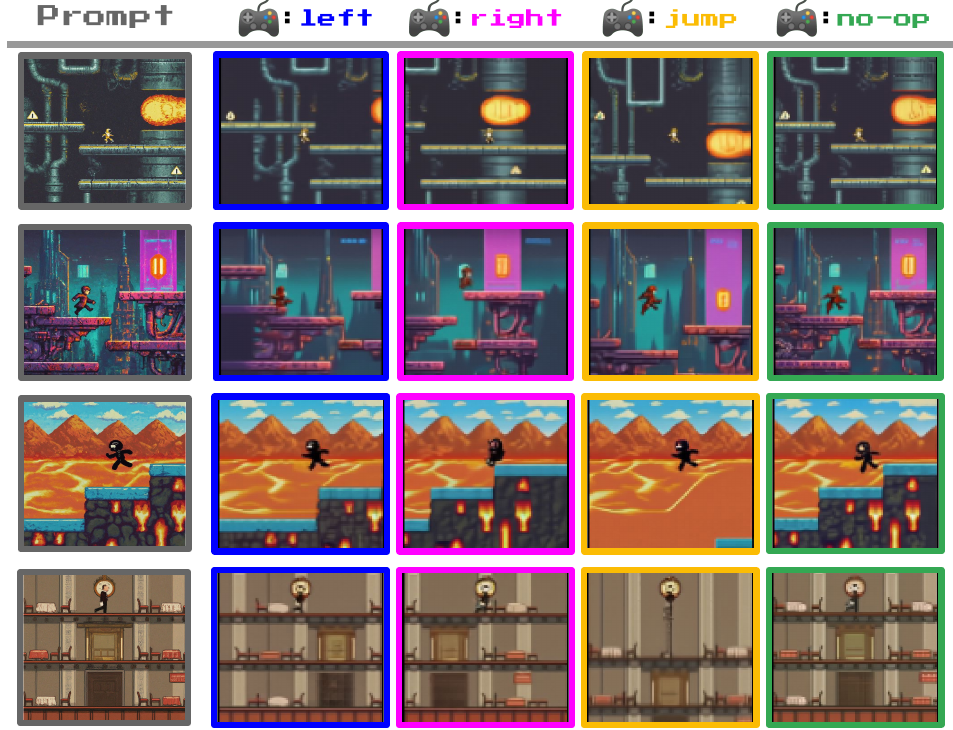}
    \caption{\textbf{Additional Simulated Game Dynamics.}
    Generated frames from~\cite{genie2024}. Each frame is generated from an initial
    synthetic prompt image from a text-to-image model, and an unsupervised latent
    action. Despite the unsupervised nature of the latent actions, their semantics
    are relatively consistent across initial frames. One limitation of the model
    is its tendency to generate relatively plain continuations outside the boundaries
    of the initial image, demonstrated most clearly in the \textbf{jump} column of
    the bottom row.
    }
    \label{fig:genie_additional_examples}
\end{figure}

The model is trained entirely unsupervised on large video datasets;
see~\ref{fig:genie_additional_examples} for example trajectories from a 10.7B
parameter model demonstrating
that the unsupervised latent action objective results in consistent control variables
across a variety of visual prompts.

\clearpage
\newpage

\section{Additional Generated Videos}
\label{app:result}

\subsection{Additional Game Simulations}

\begin{figure}[h]
    \centering
    \setlength{\tabcolsep}{-1pt}
    \begin{tabular}{cccc}
     \includegraphics[width=.2\linewidth]{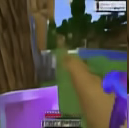} &
     \includegraphics[width=.2\linewidth]{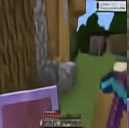} &
     \includegraphics[width=.2\linewidth]{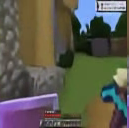} & 
     \includegraphics[width=.2\linewidth]{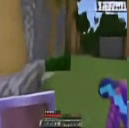} \\ 
     \includegraphics[width=.2\linewidth]{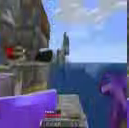} &
     \includegraphics[width=.2\linewidth]{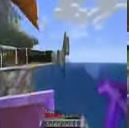} &
     \includegraphics[width=.2\linewidth]{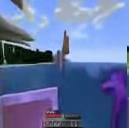} & 
     \includegraphics[width=.2\linewidth]{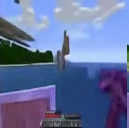} \\ 
     \includegraphics[width=.2\linewidth]{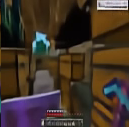} &
     \includegraphics[width=.2\linewidth]{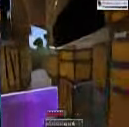} &
     \includegraphics[width=.2\linewidth]{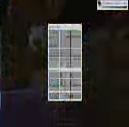} & 
     \includegraphics[width=.2\linewidth]{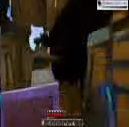} \\ 
     \includegraphics[width=.2\linewidth]{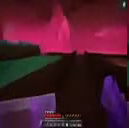} &
     \includegraphics[width=.2\linewidth]{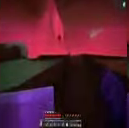} &
     \includegraphics[width=.2\linewidth]{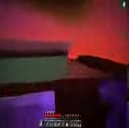} & 
     \includegraphics[width=.2\linewidth]{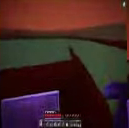} \\ 
    \end{tabular}
    \caption{\textbf{Additional Generated Game Trajectories in Minecraft.}. Using the model in Section~\ref{sec:minecraft}, we demonstrate additional rollouts from the model. We find that the model is able to handle ego-centric motion quite well. However, temporal consistency can sometimes be an issue as shown in the third row. The agent opens up an inventory mid-clip, and then the chest in front disappears. }
    \label{fig:minecraft_appendix}
\end{figure}

\clearpage
\newpage

\clearpage
\newpage

\subsection{Additional Generation for How-to Videos}

\begin{figure}[h]
    \centering
    \includegraphics[width=.8\linewidth]{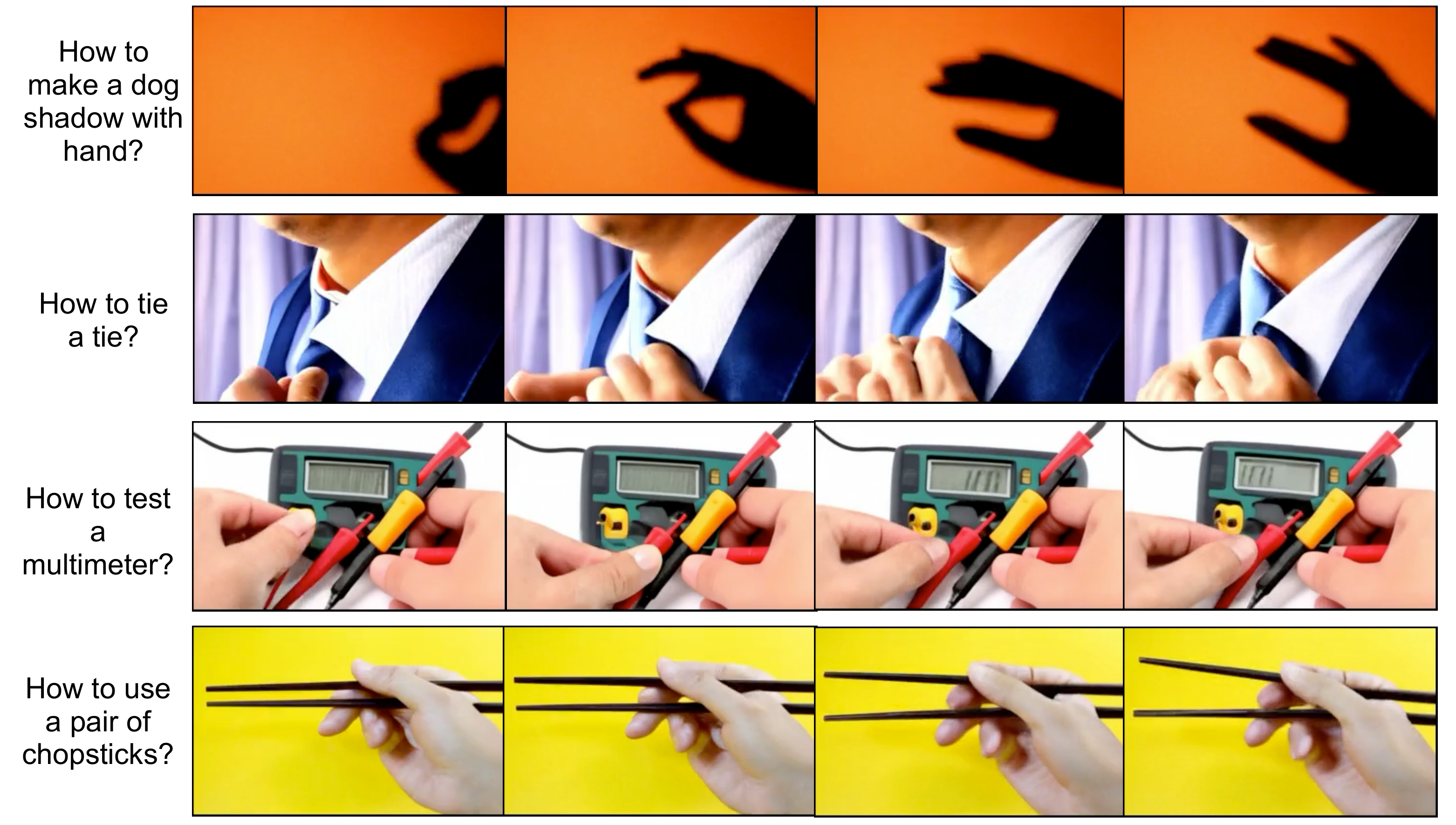}
    \caption{\textbf{Additional Generated How-to Videos.}  Some generated frames can synthesize key frames in response to human inquiries (first and last row), but some other generated frames are too generic and do not capture enough details to fully answer users' questions.}
    \label{fig:howto_app}
\end{figure}

\clearpage
\newpage

\subsection{Additional Self-Driving Simulations}

\begin{figure}[h]
    \centering
    \includegraphics[width=.8\linewidth]{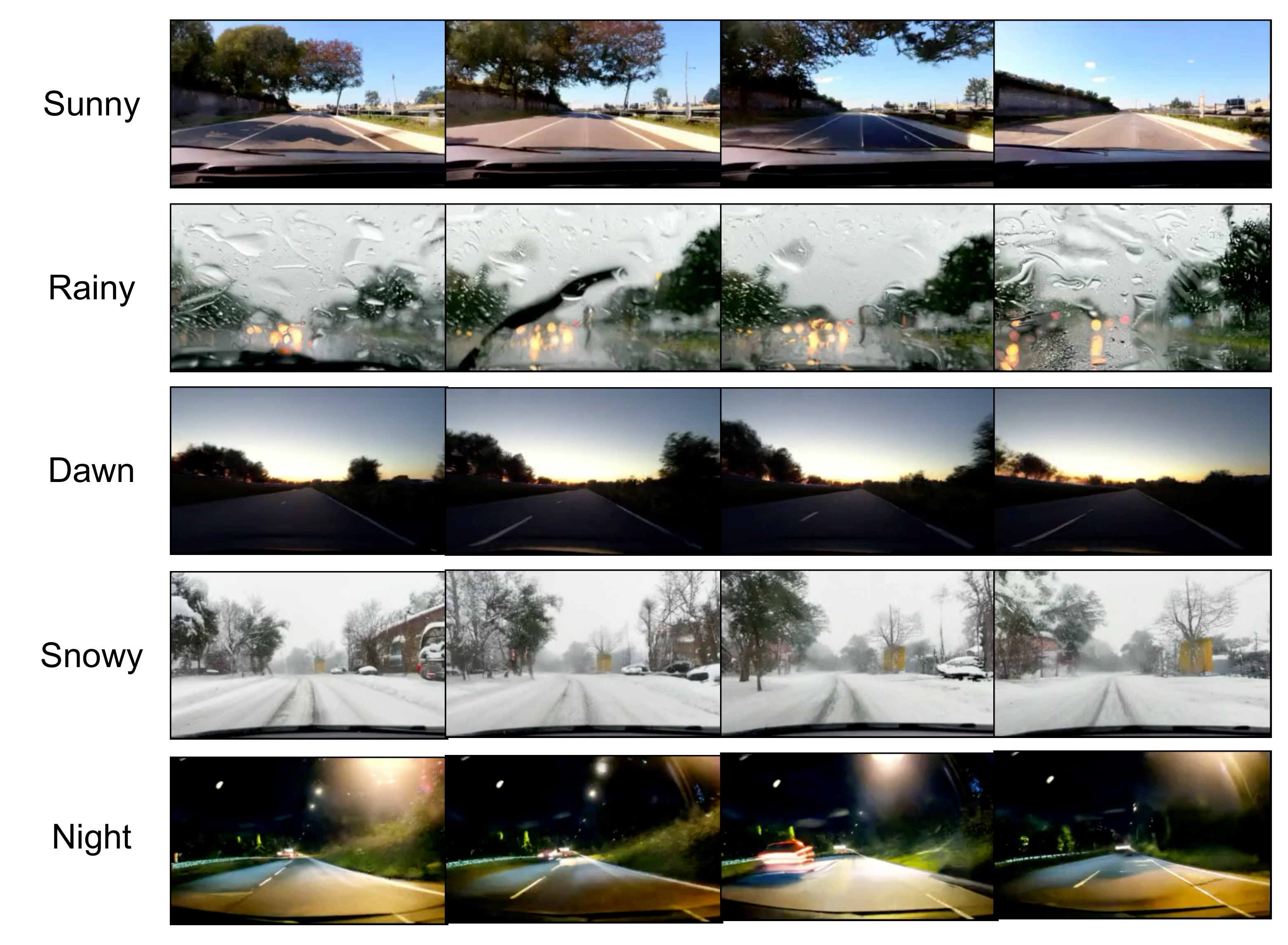}
    \caption{\textbf{Additional Generative Simulation for Driving.} Generative simulators can generate driving in different weather conditions and time of the day, such as sunny, rainy, snowy, night, and dawn.}
    \label{fig:driving_app}
\end{figure}

\clearpage
\newpage

\subsection{Additional Robot SE(3) Simulations}

\begin{figure}[h]
    \centering
    \includegraphics[width=.8\linewidth]{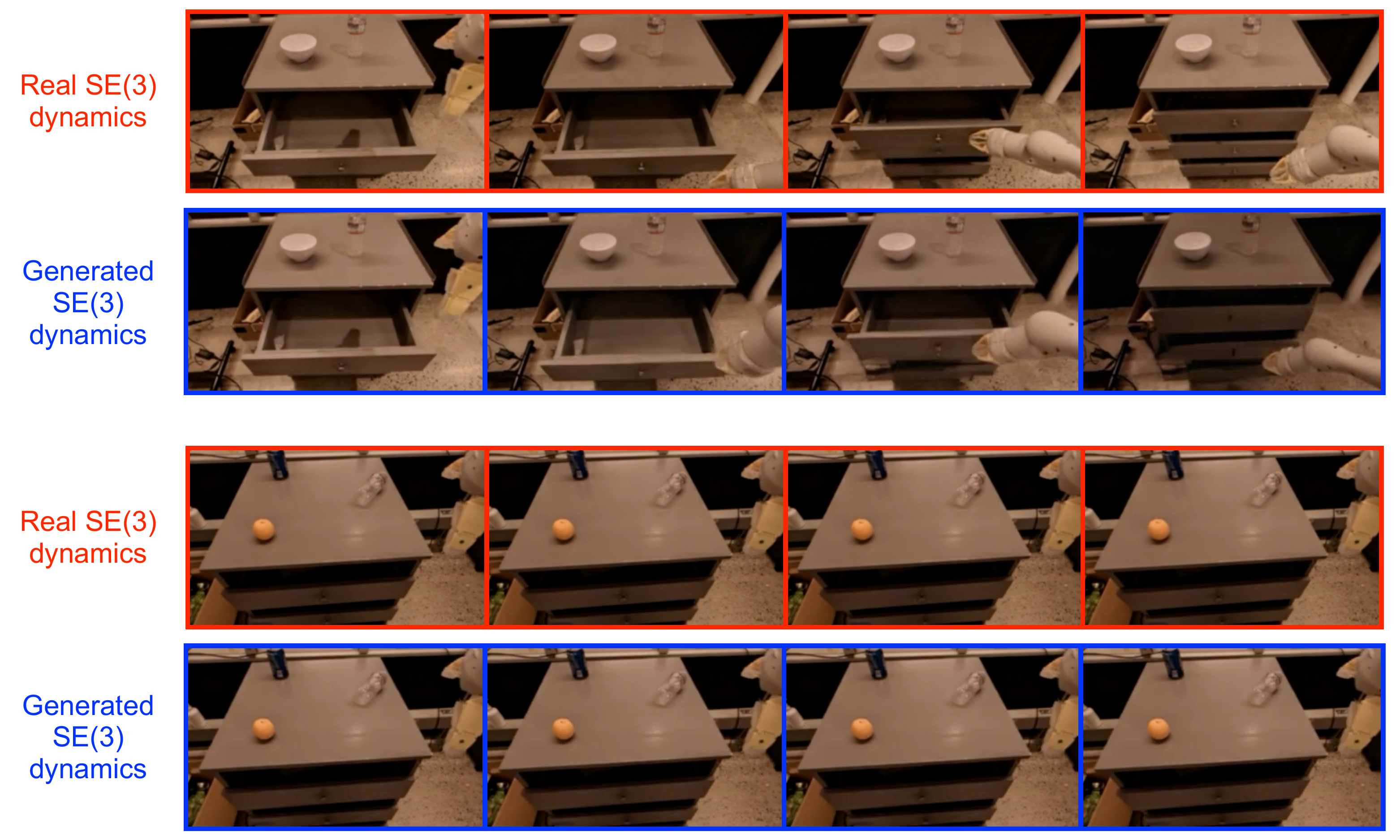}
    \caption{\textbf{Additional Generative Simulation of SE(3) Robot Actions.} Real execution of a robot policy (red) and simulated execution of the same policy (blue). The simulated rollout generally agrees with the ground truth rollout.}
    \label{fig:dynamics_app}
\end{figure}

\subsection{Examples of Hallucination}
\label{app:hallucinate}
\begin{figure}[h]
    \centering
    \includegraphics[width=.8\linewidth]{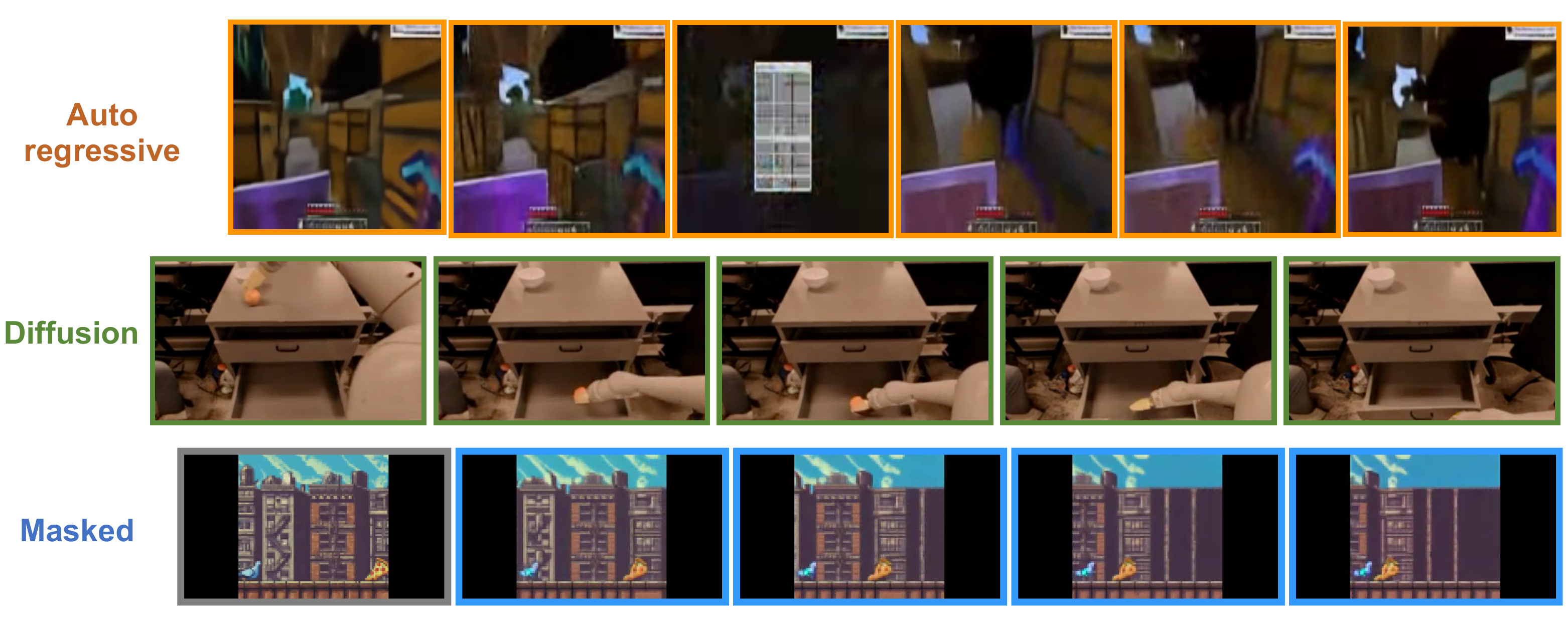}
    \caption{\textbf{Examples of Hallucination from All Three Types of Models.} The problem of hallucination persists across different types of video generation models. On the first row, the video generated by the autoregressive model shows that the chest disappears after the inventory is closed. On the second row, the video generated by the diffusion model shows the orange that the orange disappears after being put in the draw. On the bottom row, the video generated by the masked model shows that the cloud suddenly stops at the boundary.}
    \label{fig:hallucination}
\end{figure}


\end{document}